\documentclass[11pt]{article}

\usepackage{acl}

\usepackage{times}
\usepackage{latexsym}

\usepackage[T1]{fontenc}

\usepackage[utf8]{inputenc}

\usepackage{microtype}

\usepackage{inconsolata}
\usepackage{multirow}
\usepackage{amsmath}
\usepackage{amsfonts}
\usepackage{pifont}

\usepackage{graphicx}
\usepackage{algorithm}
\usepackage{algpseudocode}

\usepackage{subcaption}
\usepackage{xspace}
\usepackage{colortbl}
\usepackage{amsmath,amssymb,amsthm}

\newcommand{\ours}{Spense\xspace}
\newcommand{\oursgpt}{SpenseGPT\xspace}
\newcommand{\oursplus}{SpenseGPT+\xspace}

\usepackage{etoolbox} %

\newtoggle{showprev}

\newtoggle{shownew}
\toggletrue{shownew}

\usepackage{array} %
\makeatletter
\newcommand*{\@rowstyle}{}
\newcommand*{\rowstyle}[1]{%
  \gdef\@rowstyle{#1}%
  \@rowstyle\ignorespaces%
}
\newcolumntype{=}{%
  >{\gdef\@rowstyle{}}%
}
\newcolumntype{+}{%
  >{\@rowstyle}%
}
\makeatother

\definecolor{Emerald}{RGB}{192, 41, 66}
\iftoggle{shownew}{
    
    \newcommand{\jsleeobj}[1]{ %
    {\let\Cap\caption
    \def\caption##1{\Cap{\color{Emerald}##1}}
    \color{Emerald}#1}
    }
}
{
    
    \newcommand{\jsleeobj}[1]{#1}
    
}

\definecolor{Gray}{RGB}{200, 200, 200}
\iftoggle{showprev}{
    \newcommand{\jsleeprev}[1]{{\color{Gray}#1}}
    \newcommand{\jsleeprevobj}[1]{ %
    {\let\Cap\caption
    \def\caption##1{\Cap{\color{Gray}##1}}
    \color{Gray}#1}
    }
    \newcommand{\jsleeprevrow}[1]{\rowstyle{\color{Gray}}#1}
}
{
    \newcommand{\jsleeprev}[1]{}
    \newcommand{\jsleeprevobj}[1]{}
    \newcommand{\jsleeprevrow}[1]{}
}

\title{
\oursgpt: Practical One-shot Pruning \\Enabling Sparse and Dense GEMMs for LLM Inference
}

 \author{\textbf{
Jaeseong Lee\textsuperscript{*},
Seung-won Hwang\textsuperscript{\textdagger},
Samyam Rajbhandari\textsuperscript{*}}\\
\\
Snowflake AI Research\textsuperscript{\rm *}, Seoul National University\textsuperscript{\rm \textdagger}\\
}

\begin{document}
\maketitle
\begin{abstract}
Semi-structured 2:4 sparsity is widely supported by modern accelerators, providing up to a 2\texttimes\xspace theoretical speedup.
However, its strict 50\% sparsity constraint often causes non-negligible accuracy degradation under post-training pruning. Meanwhile, existing relaxed sparsity formats either require specialized compiler support or introduce runtime overheads that limit end-to-end speedup.
We propose \ours, a practical hybrid sparse-dense format that splits each weight matrix into a 2:4 sparse region and a dense region.
This design relaxes the effective sparsity constraint while remaining compatible with existing high-performance sparse and dense GEMM libraries, avoiding both custom compiler support and input activation expansion.
Building on this format, we introduce \oursgpt, a one-shot post-training pruning method that produces sparse and dense regions.
Notably, we show that selecting the right dense regions is important, and we devise two different strategies to choose them.
Experiments on Qwen3-32B and Seed-OSS-36B demonstrate that our method achieves up to $1.2\times$ end-to-end decoding speedup on B200 GPUs with FP8 precision, while preserving accuracy.
To the best of our knowledge, this is the first one-shot pruning demonstration of real-world end-to-end LLM decoding speedup from semi-structured sparse tensor cores on recent GPUs such as B200s, while maintaining model quality.
\end{abstract}

\section{Introduction}

Semi-structured 2:4 sparsity has emerged as a promising hardware feature for accelerating large language models (LLMs) on modern accelerators, including NVIDIA GPUs~\cite{2to4sparsityAccelerating2021mishra,B200NVIDIA}, AMD GPUs~\cite{Unlocking2026liao}, and Meta MTIAs~\cite{Meta2025}.
In this format, two out of every four consecutive weights are pruned to zero, allowing hardware to skip the corresponding load and computation, with a theoretical speedup of up to 2\texttimes~\cite{2to4sparsityAccelerating2021mishra,B200NVIDIA}.

Despite this hardware support, practical adoption of 2:4 sparsity for LLM inference remains challenging.
The main obstacle is its strict 50\% sparsity constraint: every group of four weights must retain only two nonzero values.
Maintaining model quality under this constraint often requires substantial training~\cite{Structured2023hongxiao,Sparse2025kurtic} or incurs noticeable accuracy degradation~\cite{SparseGPT2023frantar,Wandasimple2024sun,PROXSPARSE2025liu,HyperPruneLearning2026sun}, especially on challenging reasoning tasks (\autoref{tab:comparison}).

Recent works mitigate the strict 50\% sparsity requirement while retaining hardware acceleration~\cite{PATCH2025hourri,SlideSparse2026shao}. However, they are \textit{impractical}, either requiring specialized compiler support~\cite{PATCH2025hourri} or introducing runtime overheads that limit end-to-end speedup~\cite{SlideSparse2026shao}.

In this paper, we propose \ours, a \textit{practical} hybrid sparse-dense format for relaxing 2:4 sparsity. By \textit{practicality}, we mean compatibility with existing high-performance sparse and dense GEMM libraries, without requiring custom compilers or expanded activations as in existing works. Specifically, \ours splits a weight matrix into two contiguous regions: one region is pruned with the standard 2:4 sparsity pattern, while the other is kept dense ($W^T$ in \autoref{fig:impl_spense}).
This simple design has an important practical advantage: it can be executed using existing highly optimized sparse and dense GEMM libraries, such as cuSPARSELt and cuBLAS on NVIDIA GPUs, without requiring specialized compiler support.
At the same time, it avoids the runtime overhead that limits prior relaxed sparsity approaches (\autoref{tab:comparison_between_hybrid_methods}).

\begin{table*}[]
\centering
\begin{tabular}{l|cc}
\hline
            & \begin{tabular}[c]{@{}c@{}}Can readily use optimized GEMM kernels \\ (e.g., cuSPARSELt for B200)\end{tabular} & Negligible overhead \\ \hline
PATCH~\cite{PATCH2025hourri}       &                                                                                                          & \ding{51}           \\
SlideSparse~\cite{SlideSparse2026shao} & \ding{51}                                                                                                &                     \\
\ours   (ours)    & \ding{51}                                                                                                & \ding{51}           \\ \hline
\end{tabular}
\caption{Comparison of different hybrid sparse-dense methods. \ours can readily use optimized GEMM kernels while incurring negligible overhead.}
\label{tab:comparison_between_hybrid_methods}
\end{table*}

\begin{figure}[]
\centering
\includegraphics[width=\columnwidth]{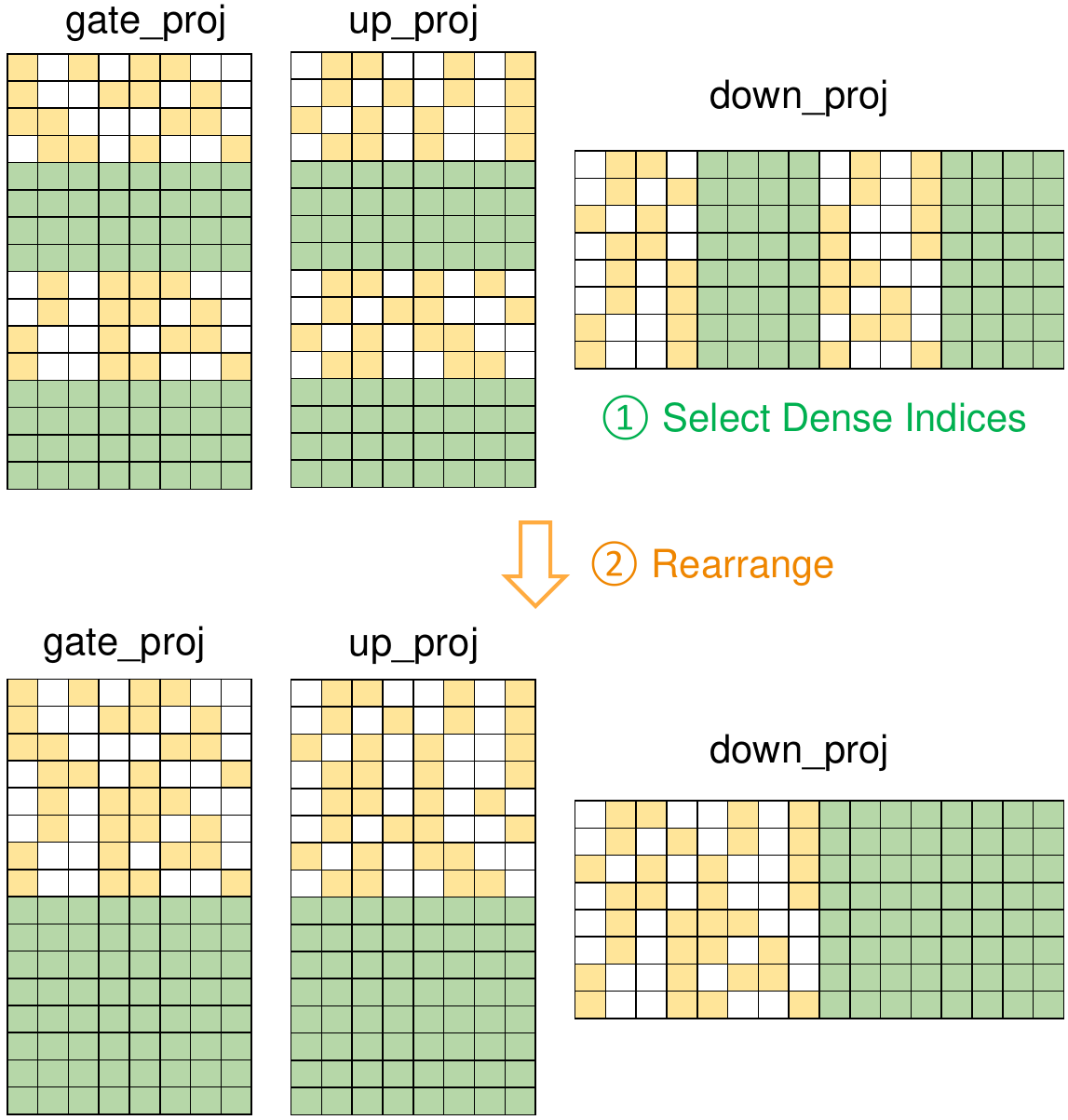}
\caption{When pruning, we select which intermediate indices to keep dense. When deploying, we rearrange the indices for kernel efficiency.}
\label{fig:impl_spensegpt}
\end{figure}

Furthermore, we propose one-shot pruning targeting \ours format.
During the pruning, we can choose arbitrary intermediate indices of GLU-style MLPs~\cite{GeGLUGLU2020shazeer} for the dense region, where each index corresponds to a row in the gate and up projections and a column in the down projection.
This is becuase intermediate indices can be permuted without changing the MLP output, yielding the contiguous layout required by our format (\autoref{fig:impl_spensegpt}).
A key question is which intermediate indices should be kept dense. We find that dense-index selection significantly impacts accuracy: even if we allow the same number of dense indices, the AIME score can plummet from 76.67 to 8.89.

We propose two methods to choose the dense indices.
First, we show that simply keeping the first $p\%$ of intermediate indices dense and applying SparseGPT~\cite{SparseGPT2023frantar} to prune the remaining regions is surprisingly good. We name this \oursgpt.
Second, we improve this through index selection based on estimated reconstruction loss.
Inspired by Wanda~\cite{Wandasimple2024sun}, we compute an activation-aware importance score for each intermediate index by combining its contributions to the gate, up, and down projections. We name this \oursplus.

We evaluate our method on Qwen3-32B~\cite{Qwen32025yang} and Seed-OSS-36B~\cite{seed2025seed-oss}.
Notably, \oursplus preserves model quality across diverse benchmarks, including AIME, GPQA, LiveCodeBench, and IFEval, with up to $1.2\times$ end-to-end \textsc{vllm} decoding speedup over the dense baseline on B200 GPUs using FP8 precision. To the best of our knowledge, this is the first one-shot pruning demonstration of real-world speedup with sparse GEMMs on recent GPUs such as B200s, while preserving model accuracy.
Our contributions are summarized as follows:
\begin{itemize}
    \item We propose \ours, a practical hybrid sparse-dense format that relaxes strict 2:4 sparsity while using existing dense and sparse GEMM libraries.
    \item We show that choosing the right dense indices is important when pruning an LLM into the \ours format.
    \item We introduce two strategies for choosing which dense indices to keep before pruning.
    \item We demonstrate end-to-end decoding speedups on B200 GPUs with FP8 precision while preserving accuracy on reasoning, instruction-following, and code-generation benchmarks.
\end{itemize}

\section{Related Work}
\subsection{Semi-Structured Sparsity in LLM Pruning}
LLM pruning can be classified into unstructured, structured, and semi-structured methods~\cite{Pruning2021behnke}. Unstructured pruning finds mask tensors that sparsify model weights~\cite{SparseGPT2023frantar,GBLMSize2024das,PrunerZero2024dong,OWLOutlier2024yin}. However, hardware acceleration is limited. Structured pruning restricts the sparsification pattern, such as removing rows, columns, or entire weight tensors, to enable more efficient hardware acceleration~\cite{LLMPruner2023ma,MINILLM2024cheng,Everybody2024dery}. Without fine-tuning, however, these methods are more vulnerable to accuracy drops than unstructured pruning~\cite{STUN2025lee}.

Semi-structured pruning, such as 2:4 sparsity, strikes a balance between the two, enabling significant speedup with a moderate accuracy drop~\cite{2to4sparsityAccelerating2021mishra,SparseGPT2023frantar,Wandasimple2024sun,PROXSPARSE2025liu,HyperPruneLearning2026sun}. This balanced structure has been introduced in various hardware platforms, such as NVIDIA GPUs~\cite{2to4sparsityAccelerating2021mishra,B200NVIDIA}, AMD GPUs~\cite{Unlocking2026liao}, and Meta MTIAs~\cite{Meta2025}.

However, the strict 50\% sparsity constraint poses challenges for practical adoption. Successful adoption requires a significant amount of training~\cite{Structured2023hongxiao,Sparse2025kurtic}; otherwise, the accuracy drop is noticeable~\cite{SparseGPT2023frantar,Wandasimple2024sun,PROXSPARSE2025liu,HyperPruneLearning2026sun}.

\subsection{Hybrid Semi-Structured Sparsity}
To mitigate the strict 50\% sparsity constraint, recent efforts have relaxed the sparsity constraint, but they remain impractical for modern GPUs such as B200. PATCH~\cite{PATCH2025hourri} proposes using dense GEMM for some tiles and sparse GEMM for other tiles, thus enabling more flexible sparsity patterns. However, to enable this, specialized Triton compiler support is required, which is not yet widely available across hardware platforms.
SlideSparse~\cite{SlideSparse2026shao} proposes expanding the input activations by $\frac{2n-2}{n}$ times to enable $(2n-2):2n$ sparsity of weights, which are expanded to the 2:4 format and compressed via existing libraries, such as cuSPARSELt. However, its speedup is limited-- for example, under a 25\% sparsity constraint, the end-to-end decoding speedup is only up to 1.04$\times$ on B200 GPUs~\cite{SlideSparse2026shao} using FP8 precision.

In contrast, \ours leverages existing highly optimized sparse and dense libraries for emerging hardware, without requiring any specialized compiler support. We also avoid input activation expansion, enabling much higher speedup than SlideSparse~\cite{SlideSparse2026shao}, as depicted in \autoref{fig:speedup}. The distinction is summarized in \autoref{tab:comparison_between_hybrid_methods}.

\section{Proposed Method}\label{sec:method}

\subsection{Preliminaries}
\subsubsection{2:4 Semi-structured sparsity}
2:4 semi-structured sparsity is a special type of sparsity pattern in which, for every 4 consecutive weights, exactly 2 are pruned to zero.
This pattern allows for efficient hardware acceleration, with up to a 2\texttimes\xspace speedup.
Formally, for a weight matrix $W \in \mathbb{R}^{N \times K}$,\footnote{We follow the format used by PyTorch, which does matrix multiplication by $XW^T$.} and input $X \in \mathbb{R}^{M \times K}$, 2:4 sparsity forces
\begin{equation}
\| W_{i,4j:4j+4} \|_0 \le 2, \forall i \le N, j \le \frac{K}{4}
\end{equation}
where $\|\cdot\|_0$ is the $L_0$ norm, which counts the number of non-zero elements.

This $W$ can be stored in a compressed format, with only non-zero weights and their positions. Since there are only a few possible ways to choose 2 non-zero weights among 4, the position metadata can be stored compactly within a few bits per group~\cite{2to4sparsityAccelerating2021mishra}. This drastically reduces the memory bandwidth requirement and, with appropriate hardware support, allows hardware to skip computations involving zero weights, leading to up to a 2\texttimes\xspace speedup. Eventually, this led to the adoption of 2:4 sparsity in various hardware platforms, such as NVIDIA GPUs~\cite{2to4sparsityAccelerating2021mishra,B200NVIDIA}, AMD GPUs~\cite{Unlocking2026liao}, and Meta MTIAs~\cite{Meta2025}.

\begin{figure}[]
\centering
\includegraphics[width=\columnwidth]{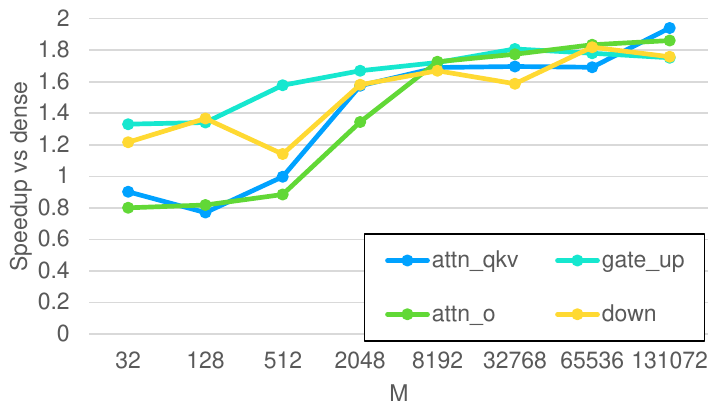}
\caption{Sparse GEMM speedup versus dense GEMM for various $M$ values on Qwen3-32B running on a B200.}
\label{fig:sparsegemm_vs_densegemm}
\end{figure}

In reality, to meet the desired speedup, we need large $M$ to amortize the overhead of sparse kernels. \autoref{fig:sparsegemm_vs_densegemm} shows the speedup of sparse GEMM versus dense GEMM for various $M$ values on Qwen3-32B running on a B200 GPU. We can see that when $M$ is small, the speedup is limited, sometimes even worse than dense GEMM, while when $M$ is large, the speedup can be up to 2\texttimes.

\subsubsection{SparseGPT}

\begin{figure}[]
\centering
\includegraphics[width=0.9\columnwidth]{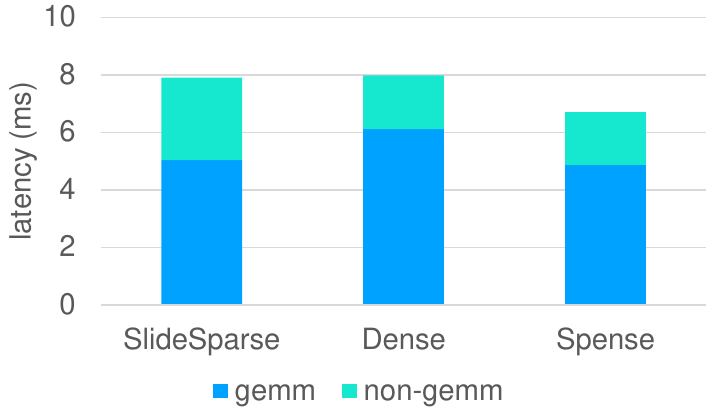}
\caption{Latency breakdown of GEMM and non-GEMM operations at $M=65536$ for the gate/up projection of Qwen3-32B on a B200. Note that PATCH~\cite{PATCH2025hourri} is not included, since it only supports Ampere GPUs.}
\label{fig:nongemm_comparison}
\end{figure}

\begin{figure*}[]
\centering
\includegraphics[width=0.9\textwidth]{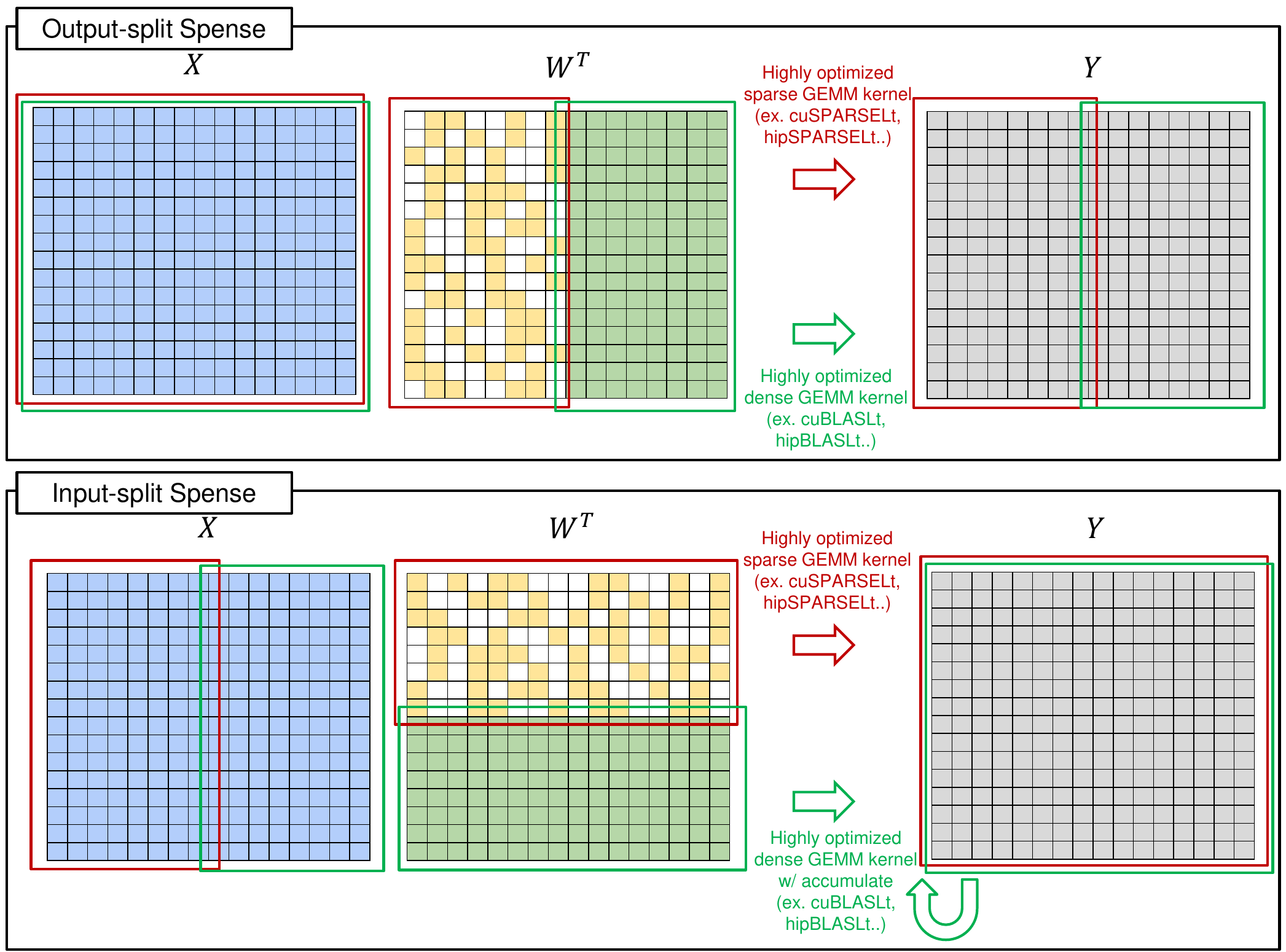}
\caption{Implementation of output-split and input-split \ours.}
\label{fig:impl_spense}
\end{figure*}

SparseGPT~\cite{SparseGPT2023frantar} is a post-training pruning method that compresses each linear layer by minimizing its output reconstruction error on calibration data.
For a weight matrix $W \in \mathbb{R}^{N \times K}$ and input activations $X \in \mathbb{R}^{M \times K}$, SparseGPT aims to find a compressed weight $\widehat{W}$ that minimizes
\begin{equation}
    \| XW^T - X\widehat{W}^T \|_F^2
\end{equation}

SparseGPT uses a second-order approximation of the layer-wise reconstruction objective.
Let $H = X^T X$ be the input Hessian approximation.
When pruning a weight $w_m$, the Optimal Brain Surgeon (OBS;~\citealp{Second1992hassibi}) update gives the compensation direction $\delta_m$ and the corresponding reconstruction error $\varepsilon_m$ as
\begin{equation}
    \delta_m
    =
    -
    \frac{w_m}{[H^{-1}]_{mm}} H^{-1}_{:,m},
    ~~
    \varepsilon_m
    =
    \frac{w_m^2}{[H^{-1}]_{mm}} 
\end{equation}
Thus, weights with smaller $\varepsilon_m$ are preferred for pruning, and the remaining weights are updated according to $\delta_m$ to reduce the reconstruction loss.

In practice, SparseGPT applies this idea column by column.
Once a column is processed, its values are frozen.
The induced error is then compensated by updating only the columns that have not yet been processed.
This column-wise ordering enables SparseGPT to reuse shared inverse-Hessian information across rows, avoiding a separate Hessian inverse for each row while still allowing different rows to have different pruning masks.

For 2:4 sparsity, within each group, SparseGPT selects the weights to prune using the OBS-based error criterion.

\subsubsection{Problems of Existing Hybrid Sparse-Dense Methods}
While 2:4 sparsity can provide significant speedup, the strict 50\% sparsity constraint poses challenges for practical adoption (\autoref{tab:comparison}).
To alleviate this issue, some recent works~\cite{PATCH2025hourri,SlideSparse2026shao} have proposed relaxing the 2:4 sparsity constraint, allowing more flexible sparsity patterns while still maintaining efficient hardware acceleration.
PATCH~\cite{PATCH2025hourri} proposes using dense GEMM for some tiles while using sparse GEMM for other tiles, enabling more flexible sparsity patterns.
SlideSparse~\cite{SlideSparse2026shao} points out that any $(2n-2):2n$ sparsity pattern is equivalent to $(n-1)$ 2:4 sparsity patterns. For example, to compress $\{a_1, a_2, a_3, a_4, 0, 0\}$, they expand the weight to $\{a_1, a_2, 0, 0, a_3, a_4, 0, 0\}$ and use standard libraries to compress the weight. To use this weight at runtime, they expand the input activation by $\frac{2n-2}{n}$ times and then run the computation with standard libraries, such as cuSPARSELt.

However, these methods have their own limitations. PATCH~\cite{PATCH2025hourri} requires specialized Triton compiler support, which is implemented only for Ampere GPUs at the time of writing.\footnote{\url{https://github.com/Paramathic/stoicc/blob/602676325280316dfbd12449df0c8d73e15f5b98/README.md}} SlideSparse~\cite{SlideSparse2026shao} requires input activation expansion, which leads to limited speedup, especially when $M$ is large (\autoref{fig:nongemm_comparison}), where sparse GEMM can maximize its strength (\autoref{fig:sparsegemm_vs_densegemm}).

\subsection{\ours: Practical Hybrid of Sparse and Dense}
Instead, we propose \ours, which avoids the limitations of previous methods while still allowing flexible sparsity patterns. \ours splits the weight matrix into two parts: one part, covering $p$\%, is kept dense, and the other $100-p$\% is pruned with the 2:4 sparsity pattern (\autoref{fig:impl_spense}). This allows us to leverage the existing highly optimized sparse and dense libraries for emerging hardware, such as cuSPARSELt and cuBLASLt for NVIDIA GPUs, or hipSPARSELt and hipBLASLt for AMD GPUs.
This also avoids input activation expansion, enabling much higher speedup than SlideSparse~\cite{SlideSparse2026shao}, as depicted in \autoref{fig:speedup}.

We implement two efficient variants of \ours. \textit{Output-split} \ours splits the output tensor into two parts and uses a sparse GEMM kernel for one part and a dense GEMM kernel for the other part. \textit{Input-split} \ours splits the input tensor into two parts. First, it uses a sparse GEMM kernel for the first part. Then it reuses the output tensor to call a dense GEMM kernel, with accumulation enabled.
For GLU variants in recent LLMs~\cite{GeGLUGLU2020shazeer}, we use the output-split format for $W_{up}, W_{gate}$, and the input-split format for $W_{down}$.

\subsection{\oursgpt and \oursplus}

\subsubsection{Choosing Which Indices to Keep Dense}

The direct \ours format splits a weight matrix into two contiguous regions: one dense region and one 2:4 sparse region.
This direct format is efficient because it avoids scattered indexing.

However, keeping this contiguous region even at the pruning stage is unnecessarily restrictive.
Given any permutation matrix $P$ over the intermediate dimension, the dense MLP output is unchanged under
\begin{multline}
\text{MLP}(X)=
\left(
X W_{up}^T
\odot
\sigma(X W_{gate}^T)
\right)
W_{down}^T
\\=
(
X(P^T W_{up})^T
\odot \\
\sigma(X(P^T W_{gate})^T)
)
(W_{down}P)^T 
\end{multline}

Throughout this section, an intermediate index denotes one coordinate of the MLP intermediate dimension.
Each such index corresponds to a row in $W_{gate}$ and $W_{up}$ and the matching column in $W_{down}$.
Thus, we can first choose which intermediate indices should be dense, and then permute those indices into a contiguous region for efficient execution.
This gives \ours a higher degree of freedom than the direct format, while preserving the same runtime structure.

We find that \textit{choosing the right dense indices is highly important}. For example, \autoref{tab:dense_selection} shows that some dense-index sets can significantly degrade performance. We identify two strategies for selecting dense indices as follows.

\subsubsection{\oursgpt: Simple and Effective Dense-Index Selection}
Our first strategy is simply to select the last contiguous region, as depicted in \autoref{fig:impl_spense}.
We choose the last $p\%$ of indices, rather than the first $p\%$, because of the column-wise execution of SparseGPT.
SparseGPT processes columns from left to right, freezes the processed columns, and compensates for their reconstruction error by updating only the columns that remain to the right.
Therefore, placing dense columns on the right keeps them available as a compensation buffer while the sparse columns are processed.
Formally, given a dense ratio $p$ and a number of intermediate indices $N$, we select
\begin{equation}
    \mathcal{D}_{\text{\oursgpt}}
    =
    \{\lfloor pN \rfloor,\dots,N\}.
\end{equation}

This simple strategy works surprisingly well, as we show in Section \ref{sec:exp}.

\subsubsection{\oursplus: Dense-Index Selection by Estimating Reconstruction Loss}

We further aim to select dense indices that are expected to contribute most to reconstruction quality.
Given a dense-index set $\mathcal{D}$, rows indexed by $\mathcal{D}$ are kept dense in $W_{gate}$ and $W_{up}$, and columns indexed by $\mathcal{D}$ are kept dense in $W_{down}$.
Ideally, we would choose $\mathcal{D}$ to minimize the MLP reconstruction loss:
\begin{equation}
\min_{\mathcal{D}: |\mathcal{D}| = pN}
\big\|
X W_{mlp}^T-X \widehat{W}_{mlp}(\mathcal{D})^T
\big\|_F^2
\end{equation}
where $\widehat{W}_{mlp}(\mathcal{D})$ denotes the compressed weights under the shared dense-index set $\mathcal{D}$, for $mlp \in \{up, gate, down\}$.
However, directly optimizing this objective is combinatorial and impractical.

We therefore use a one-shot loss estimator to assign an importance score to each intermediate index.
The desired estimator should be independent across indices, so that we can greedily keep the indices with the largest estimated reconstruction error.
We instantiate this estimator with a Wanda-style score~\cite{Wandasimple2024sun}, which estimates weight importance using both weight magnitude and input activation magnitude.
For a linear layer with weight $W \in \mathbb{R}^{N \times K}$ and input activations $X$, removing weight $W_{ij}$ changes the output by $W_{ij}X_{:,j}$.
Thus, its squared contribution to the reconstruction error is proportional to
\begin{equation}
    W_{ij}^2 \|X_{:,j}\|_2^2 .
\end{equation}
Using the input Hessian approximation $H=X^TX$, we have $\|X_{:,j}\|_2^2 = H_{j,j}$, giving the element-wise importance score
\begin{equation}
    S_{ij}
    =
    |W_{ij}|\sqrt{H_{j,j}} .
\end{equation}
Since $S_{ij}^2$ approximates the reconstruction error from removing $W_{ij}$, aggregating $S_{ij}$ over a row or column provides a simple estimate of the importance of that row or column.

For the GLU-variant MLP triplet~\cite{GeGLUGLU2020shazeer}, which is prevalent in recent LLM architectures~\cite{Qwen32025yang,seed2025seed-oss}, each intermediate index appears as a row in $W_{gate}$ and $W_{up}$, and as a column in $W_{down}$.
We therefore estimate the importance of index $i$ by combining all three contributions. For $mlp \in \{up, gate, down\}$, we define importance as:
\begin{equation}
    s^{mlp}_i=
    \frac{1}{K}
    \sum_{j=1}^{K}
    |(W_{mlp})_{i,j}|
    \sqrt{(H_{mlp})_{j,j}}
\end{equation}

Because the three projections can have different score scales, we normalize each score vector by its mean:
\begin{equation}
    \operatorname{Norm}(s)_i
    =
    \frac{s_i}{\frac{1}{N}\sum_{k=1}^{N}s_k+\epsilon}.
\end{equation}
The final estimated reconstruction-loss priority for index $i$ is
\begin{equation}
    s_i
    =
    \operatorname{Norm}(s^{gate})_i
    +
    \operatorname{Norm}(s^{up})_i
    +
    \operatorname{Norm}(s^{down})_i .
\end{equation}
Indices with larger $s_i$ are estimated to induce larger reconstruction loss if compressed, and are therefore prioritized to remain dense.

We then select the dense-index set by taking the top-scoring indices under the required alignment constraint:
\begin{equation}
    \mathcal{D}_{\text{\oursplus}}
    =
    \operatorname{TopK}_{pN}(s),
\end{equation}
where $\operatorname{TopK}_{pN}$ denotes selecting approximately $pN$ indices with the largest scores.

After selecting $\mathcal{D}_{\text{\oursplus}}$, we permute the intermediate dimension so that the selected dense indices become contiguous.
We then apply SparseGPT~\cite{SparseGPT2023frantar} to the sparse regions while protecting the selected dense indices.
As in \oursgpt, we place the selected dense indices in the lower/right part before pruning, following the same SparseGPT compensation rationale.

\begin{table*}[]
\centering
\setlength{\tabcolsep}{2pt}
\begin{tabular}{lc|ccccc|ccccc}
\hline
           &              & \multicolumn{5}{c|}{Qwen3-32B}                                                                                         & \multicolumn{5}{c}{Seed-OSS-36B-Instruct}                                                                              \\ \hline
           & MLP Sparsity & AIME           & GPQA           & IFEval         & \multicolumn{1}{c|}{LCB}                           & avg            & AIME           & GPQA           & IFEval         & \multicolumn{1}{c|}{LCB}                           & avg            \\ \hline
\rowcolor[HTML]{EFEFEF} 
Original   & 0\%           & 75.56          & 65.49          & 85.46          & \multicolumn{1}{c|}{\cellcolor[HTML]{EFEFEF}60.19} & 71.68          & 83.33          & 69.19          & 81.95          & \multicolumn{1}{c|}{\cellcolor[HTML]{EFEFEF}60.95} & 73.86          \\
SparseGPT  & 50\%          & 68.89          & 53.71          & 75.42          & \multicolumn{1}{c|}{43.05}                         & 60.27          & 75.56          & 60.27          & 80.90          & \multicolumn{1}{c|}{44.00}                         & 65.18          \\
HyperPrune & 50\%          & 13.33          & 23.57          & 54.65          & \multicolumn{1}{c|}{5.33}                          & 24.22          & 1.11           & 7.58           & 33.77          & \multicolumn{1}{c|}{0.38}                          & 10.71          \\
Wanda      & 50\%          & 6.67           & 15.49          & 49.85          & \multicolumn{1}{c|}{3.81}                          & 18.96          & 0.00           & 6.90           & 31.67          & \multicolumn{1}{c|}{0.95}                          & 9.88           \\
\oursgpt   & 37.5\%        & 73.33          & 59.43          & 77.14          & \multicolumn{1}{c|}{51.62}                         & 65.38          & 78.89          & 64.31          & 82.75          & \multicolumn{1}{c|}{55.04}                         & 70.25          \\
\oursplus  & 37.5\%        & 77.78          & 61.45          & 79.36          & \multicolumn{1}{c|}{52.57}                         & 67.79          & 76.67          & 66.50          & 83.73          & \multicolumn{1}{c|}{54.48}                         & 70.35          \\
\oursgpt   & 25\%          & 71.11          & 62.46          & 80.78 & \multicolumn{1}{c|}{54.48}                         & 67.21          & 77.78          & 66.84          & 82.50          & \multicolumn{1}{c|}{59.62}                         & 71.69          \\
\oursplus  & 25\%          & \textbf{76.67} & \textbf{66.83} & \textbf{81.83}          & \multicolumn{1}{c|}{\textbf{57.71}}                & \textbf{70.76} & \textbf{80.00} & \textbf{68.18} & \textbf{83.73} & \multicolumn{1}{c|}{\textbf{60.57}}                & \textbf{73.12} \\ \hline
\end{tabular}
\caption{Evaluation results on Qwen3-32B and Seed-OSS-36B. We average three runs with different random seeds for each method.}
\label{tab:comparison}
\end{table*}

\section{Experiments}\label{sec:exp}
\subsection{Experimental Settings}
\paragraph{Target LLMs}
We evaluate on two different families of LLMs, Qwen3-32B~\cite{Qwen32025yang} and Seed-OSS-36B~\cite{seed2025seed-oss}.

\paragraph{Datasets}
For calibration, we use 128 samples from s1K-1.1~\cite{s12025muennighoff}, with a sequence length of 16384.
For evaluation, we use a diverse set of benchmarks covering mathematical reasoning (AIME2024), knowledge-intensive question answering (GPQA Diamond), instruction following (IFEval), and code generation (LiveCodeBench v6; LCB).

\paragraph{Experimental Details}
We evaluate the end-to-end decoding speedup on B200 GPUs using FP8 precision and \textsc{vllm}~\cite{vllmEfficient2023kwon} as the decoding framework, following previous work~\cite{SlideSparse2026shao}. We use a batch size of 32 and a sequence length of 4096.

We use \textsc{evalscope}~\cite{evalscope_2024}\footnote{We used evalscope instead of lm-evaluation-harness or lighteval, since some tasks were not supported or the task performance of the original Qwen3-32B was much lower.} to evaluate the task performance. We use a maximum of 32768 new tokens. For Qwen3-32B, we use a temperature of 0.6 and a top-p of 0.95. For Seed-OSS-36B, we use a temperature of 1.1, a top-p of 0.95, and a thinking budget of 16384. We run each experiment three times and report the average.

When pruning, we focus on MLPs, keeping attention parameters as-is. We use true-sequential mode when we apply SparseGPT.

\begin{figure}[]
\centering
\includegraphics[width=0.9\columnwidth]{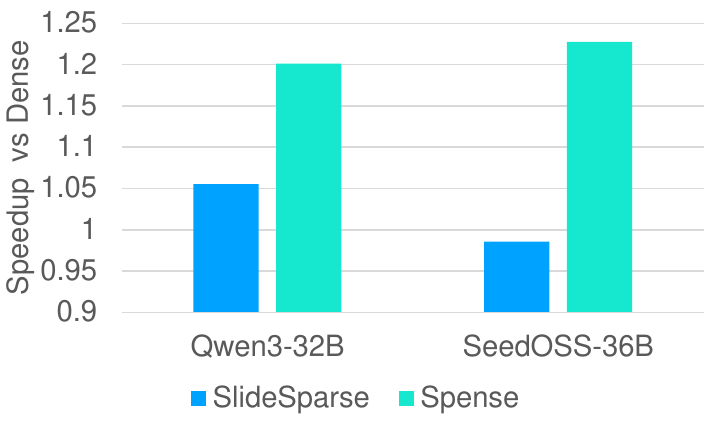}
\caption{End-to-end decoding speedup over the dense baseline, targeting 25\% sparsity. We compare \ours with SlideSparse on B200 GPUs using FP8 precision and \textsc{vllm}.}
\label{fig:speedup}
\end{figure}

\subsection{Experimental Results}

\subsubsection{Accuracy Preservation}

\autoref{tab:comparison} compares the accuracy of strict 2:4 pruning baselines and our hybrid sparse-dense methods.
Across both Qwen3-32B and Seed-OSS-36B, strict 2:4 pruning substantially degrades accuracy on challenging benchmarks.
Taking Seed-OSS-36B as an example, SparseGPT at 50\% MLP sparsity reduces the average score from 73.86 to 65.18, while Wanda and HyperPrune degrade further.

In contrast, \oursgpt and \oursplus preserve accuracy much more effectively.
On Seed-OSS-36B, \oursgpt at 25\% MLP sparsity reaches 71.69, while \oursplus reaches 73.12, nearly matching the dense baseline average of 73.86.
On Qwen3-32B, \oursgpt at 25\% MLP sparsity reaches an average score of 67.21, and \oursplus further improves it to 70.76, close to the dense baseline score of 71.68.
These results show that relaxing strict 2:4 sparsity through a dense region is critical for preserving model quality under post-training pruning.

The comparison between \oursgpt and \oursplus also shows that dense-index selection matters: on Qwen3-32B, \oursplus at 37.5\% MLP sparsity reaches a 67.79 average score, even comparable to \oursgpt at 25\% sparsity.
On the same sparsity, \oursplus consistently improves over \oursgpt by selecting dense indices using the proposed reconstruction-loss estimator.
This confirms that our proposed reconstruction-loss estimator is effective for selecting important dense indices.

\subsubsection{Real-world End-to-End Speedup}
\autoref{fig:speedup} reports the end-to-end decoding speedup over the dense baseline.
\ours achieves up to 1.2\texttimes\xspace end-to-end speedup, demonstrating that the proposed sparse-dense format can translate sparse tensor core acceleration into real serving-level gains, while SlideSparse~\cite{SlideSparse2026shao} does not.
\autoref{fig:nongemm_comparison} details the reason. SlideSparse~\cite{SlideSparse2026shao} can reduce GEMM time, but its additional non-GEMM overhead prevents it from reducing the total latency of the linear layer.
In contrast, \ours retains the benefit of sparse tensor cores while keeping non-GEMM overhead small.

\begin{table}[]
\centering
\begin{tabular}{l|cc}
\hline
           & Qwen3-32B & Seed-OSS-36B \\ \hline
SparseGPT  & 1.33h      & 1.52h         \\
HyperPrune & 3.80h      & 4.07h         \\
ProxSparse & OOM      & OOM         \\
\oursgpt   & 1.33h      & 1.52h         \\
\oursplus  & 1.87h      & 2.15h         \\ \hline
\end{tabular}
\caption{Pruning time (in hours) of different methods on Qwen3-32B and Seed-OSS-36B on a single B200 GPU.}
\label{tab:pruning_time}
\end{table}

\subsubsection{Pruning Time Efficiency}
\autoref{tab:pruning_time} compares the pruning time on a single B200 GPU in hours. \oursgpt and \oursplus can prune 32B-36B LLMs within 2.15 hours, which is comparable to SparseGPT~\cite{Sparse2025kurtic} and much faster than HyperPrune~\cite{HyperPruneLearning2026sun}. ProxSparse~\cite{PROXSPARSE2025liu} runs out of memory when pruning these large models on a single GPU.

\begin{table}[]
\centering
\setlength{\tabcolsep}{1pt}
\begin{tabular}{l|cccc|c}
\hline
               & AIME           & GPQA           & IFEval         & LCB            & avg            \\ \hline
SparseGPT & 8.89           & 27.95          & 65.19          & 9.14           & 27.79          \\
\oursplus      & \textbf{76.67} & \textbf{66.83} & \textbf{81.83} & \textbf{57.71} & \textbf{70.76} \\ \hline
\end{tabular}
\caption{Comparison of different dense-index selection strategies on Qwen3-32B. The SparseGPT strategy selects dense rows and columns based on the SparseGPT criterion, while \oursplus selects them based on our reconstruction-loss importance score.}
\label{tab:dense_selection}
\end{table}

\begin{table}[]
\centering
\setlength{\tabcolsep}{1pt}
\begin{tabular}{l|cccc|c}
\hline
            & AIME           & GPQA           & IFEval         & LCB            & avg            \\ \hline
None        & 12.22          & 28.45          & 66.29          & 7.05           & 28.50          \\
Front       & 13.33          & 28.45          & 67.53          & 9.14           & 29.61          \\
Back (ours) & \textbf{76.67} & \textbf{66.83} & \textbf{81.83} & \textbf{57.71} & \textbf{70.76} \\ \hline
\end{tabular}
\caption{Comparison of placement strategies before applying the pruning algorithm in \oursplus on Qwen3-32B.}
\label{tab:front_back}
\end{table}

\subsubsection{Ablation on Dense-Index Scoring}

We next study whether our proposed index importance scoring is important.
Our key assumption was that an index importance score should be independent of other indices.
To break this assumption, we leverage a SparseGPT-like score as follows.
Following SparseGPT, we compute the inverse-Hessian-based saliency
\begin{equation}
    S_{ij}^{\text{SGPT}}
    =
    \frac{W_{ij}^2}{[H^{-1}]_{j,j}},
\end{equation}
and aggregate it over rows or columns to obtain an intermediate-index priority.

As shown in \autoref{tab:dense_selection}, the SparseGPT-like score performs poorly for dense-index selection, achieving an average score of only 27.79, while \oursplus achieves 70.76.
We hypothesize that this is because SparseGPT inherently assumes dependencies on other weights, requiring reconstruction to apply it properly. Designing better dense-index selection methods is left for future work.

\subsubsection{Ablation on Dense-Index Layout}

Finally, we study how the placement of selected dense indices affects the final compressed model.
After selecting dense indices, we can either keep them in place, move them to the front/upper side, or move them to the back/lower side before applying SparseGPT.

\autoref{tab:front_back} shows that the layout choice has a dramatic impact.
Keeping the selected indices in place achieves an average score of only 28.50, and moving them to the front achieves 29.61.
In contrast, moving them to the back before SparseGPT reconstruction achieves an average score of 70.76.
This validates the design choice described in Section \ref{sec:method}: SparseGPT processes columns from left to right, freezes processed columns, and compensates for reconstruction error using only the columns that remain to the right.
Placing dense columns on the right keeps them available as a compensation buffer while sparse columns are processed in $W_{down}$.

\section{Conclusion}

We presented \ours, a hybrid sparse-dense format that makes semi-structured sparsity more deployable for LLM inference.
We showed that the selection and layout of dense indices matter, and we proposed two strategies, \oursgpt and \oursplus.
Experiments on Qwen3-32B and Seed-OSS-36B demonstrate that \ours achieves real end-to-end decoding speedups on B200 GPUs with FP8 precision while preserving model quality.

\section*{Limitations}
The speedup of \ours depends on the efficiency of the underlying sparse and dense GEMM libraries.
The realized speedup can vary depending on batch size, sequence length, hardware backend, precision, and serving configuration.
While we evaluate on B200 GPUs with FP8 precision and \textsc{vllm}, additional evaluation on other accelerators and deployment stacks is needed to fully characterize portability.

\oursplus relies on calibration data to estimate dense-index importance.
Although we use a small calibration set, the selected dense indices may depend on the calibration distribution.
If the deployment distribution differs substantially from the calibration data, the selected indices may no longer be optimal.
Studying more robust or task-adaptive dense-index selection strategies is an important direction for future work.

\bibliography{2026ARR_Spense.bib,custom.bib}

\end{document}